# The MC$^2$ Project
# [Machines of Collective Conscience]

A possible walk, up to Life-like Complexity and Behaviour, from bottom, basic and simple bio-inspired heuristics – a walk, up into the *morphogenesis* of information.

**Vitorino Ramos**
[http://alfa.ist.utl.pt/~cvrm/staff/vramos]

"[…]QUESTION_HUMAN > If Control's control is absolute, why does Control need to control?
ANSWER_CONTROL > Control…, needs time.
QUESTION_HUMAN > Is Control controlled by his need to control ?
ANSWER_CONTROL > Yes.
QUESTION_HUMAN > Why is Control need Humans, has you call them ?
ANSWER_CONTROL > Wait ! Wait…! Time are lending me…;
Death needs time like a Junkie… needs Junk.
QUESTION_HUMAN > And what does Death need time for ?
ANSWER_CONTROL > The answer is so simple ! Death needs time for what it kills to grow in ! […]",

in Dead City Radio, *William S. Burroughs / John Cale* , 1990.

      Imagine uma "Máquina" onde não existe algum tipo de constrangimento a nenhum tipo particular de representação: o comportamento desejado é distríbuido e levemente especifícado simultaneamente ao longo de todas as suas componentes básicas, sem que exista um minímo de especifícação desse mesmo mecanismo necessário para gerar esse tipo de comportamento, isto é, o comportamento global emerge a partir das relações existentes entre os múltiplos e simples comportamentos. Uma máquina que vive para, e de/com a sinergia.

      A emergência de comportamento complexo num sistema composto por simples elementos interagindo entre si, é um dos mais fascinantes fenómenos da natureza, e do nosso mundo. Exemplos podem ser encontrados em práticamente todos os campos científicos contemporâneos de interesse, desde a formação coerente de padrões em sistemas químicos e físicos, passando pelo movimento de grupo de diferentes espécies de animais em biologia, até ao comportamento social de massas. Nas ciências sociais por exemplo, é usual a convicção que a evolução dos sistemas sociais é determinada por numerosos factores, difíceis de explorar, tais como aqueles que são de um modo ou de outro directamente relacionados com a cultura, a sociologia, a economia, a politica, a ecologia, etc. Contudo, em anos recentes, o desenvolvimento de campos científicos interdisciplinares tais comos as "Ciências da Complexidade" e da "Vida Artificial", conduziram à percepção que processos de dinâmica complexa podem igualmente resultar de simples interacções. Aliado a este aspecto, e para um determinado nivel de abstracção, verifica-se ainda que se podem encontrar muitas características comuns em diferentes campos de estudo.

Moreover, at a certain level of abstraction, one can also find many common features between complex structures in very different fields. For instance, it is an old idea that society is in a number of respects similar to an organism, a living system with its cells, metabolic circuits and systems. As an example, the army functions like an immune system, protecting the organism from invaders, while the government functions like the brain, steering the whole and making decisions. In this metaphor, different organizations or institutions play the role of organs, each fulfilling its particular function in keeping the system alive, an idea that can be traced back at least as far as *Aristotle*, being a major inspiration for the founding fathers of sociology, such as *Comte*, *Durkheim* and especially *Spencer*.

The organismic view of society has much less appeal to contemporary theorists. Their models of society are much more interactive, open-ended, and non-deterministic than those of earlier sociologists, and they have learned to recognize the intrinsic complexity and unpredictability of society. The static, centralized, hierarchical structure with its rigid division of labor that seems to underlie the older organismic models appears poorly suited for understanding the intricacies of our fast-evolving society. Moreover, a vision of society where individuals are merely little cells subordinated to a collective system has unpleasant connotations to the totalitarian states created in the last century. As a result, the organismic model is at present generally discredit in sociology.

Similarly, biology has traditionally started at the top, viewing a living organism as a complex biochemical machine, and has worked *analytically* down from there through the hierarchy of biological organization – decomposing a living organism into organs, tissues, cells, organelles, and finally molecules – in its pursuit of the mechanisms of life. Analysis means 'the separation of an intellectual or substantial whole into constituents for individual study' (that is, by top-down reductionist approaches). By composing our individual understandings of the dissected component parts of living organisms, traditional biology has provided us with a broad picture of the mechanics of life on Earth.

In the meantime, however, new scientific developments have done away with rigid, mechanistic views of organisms. As pointed by *Langton*, there is more to life than mechanics – there is also dynamics. Life depends critically on principles of dynamical self-organization that have remained largely untouched by traditional analytic methods. There is a simple explanation for this – these self-organized dynamics are fundamentally non-linear phenomena, and non-linear phenomena in general depend critically on the interactions between parts: they necessarily disappear when parts are treated in isolation from one another, which is the basis for any analytic method. Rather, non-linear phenomena are most appropriately treated by a *synthetic* approach, where synthesis means "the combining of separate elements or substances to form a coherent whole'. In non-linear systems, the parts must be treated in each other's presence, rather than independently from one another, because they behave very differently in each other's presence than we would expect from a study of the parts in isolation. Of course, there is no universally agreed definition of life. The concept covers a cluster of properties, most of which are themselves philosophically problematic: self-organization, emergence, autonomy, growth, development, reproduction, evolution, adaptation, responsiveness, and metabolism. Scientists differ about the relative importance of these properties, although it is generally agreed that the possession of most (not necessarily all) of them suffices for something to be regarded as alive.

Similarly, when studying living systems, biologists no longer focus on the static structures of their anatomy, but on the multitude of interacting processes that allow the organism to adapt to an ever changing environment. Recently, the variety of ideas and methods that is commonly grouped under the header of "the sciences of complexity" along with *Artificial Life*, has led to understanding that organisms are self-organizing, adaptive systems. Most processes in such systems are decentralized, non-deterministic and in constant flux. They thrive on noise, chaos and creativity. Their collective swarm-intelligence emerges out of the free interactions between individually autonomous components. Rather than take living things apart, Artificial Life attempts to put living things together within a *bottom-up* approach, that is, beyond *life-as-we-know-it* into the realm of *life-as-it-could-be*, generating *lifelike* behaviour, and focusing on the problem of creating behaviour generators, inspired on the nature itself, even if the results (what emerges from the process) have no analogues in the *natural* world. The key insight into the natural method of behaviour generation is gained by noting that nature is fundamentally parallel. This is reflected in the "architecture" of natural living organisms, which consist of many millions of parts, each one of which has its own behavioural repertoire. Living systems are highly distributed and quite massively parallel.

Artificial Intelligence (AI) and aLife are each concerned with the application of computers to the study of complex, natural phenomena. Apart from traditional and symbolic top-down AI in the sixties and seventies, both are nowadays concerned with generating complex behaviour, in a bottom-up manner, turning their attention from the *mechanics* of phenomena to the *logic* of it. The first computational approach to the generation of lifelike behaviour was due to the mathematician *John Von Neumann*. In the words of his colleague *Arthur W. Burks*, *Von Neumann* was interested in the general question:

> […] What kind of logical organization is sufficient for an automaton to reproduce itself ? This question is not precise and admits to trivial versions as well as interesting ones. *Von Neumann* had the familiar natural phenomenon of self-reproduction in mind when he posed it, but he was trying to simulate the self-reproduction of a natural system at the level of genetics and biochemistry. He wished to abstract from the natural self-reproduction problem its logical form […]

This approach is the first to capture the essence of Artificial Life (replace, for instance, references to 'self-reproduction' in the above with references to any other biological phenomena). From this "kinematic model" of *Von Neumann*, a genuine self-reproduction mechanism implemented in the sixties, *Stan Ulam* suggested an appropriate formalism where the logical form of the process is completely distinguish from the material counterpart, which has come to be know as a *Cellular Automata* (CA). In brief, a CA consists of a regular lattice of (many) *finite automata*, which are the simplest formal models of machines. A finite automata can be in only one of a finite number of states at any given time, and its transition between states from one time-step to the next are governed by a *state-transition table*: given a certain input and a certain internal state, the state-transition table specifies the state to be adopted by the finite automata at the next time step. In a CA, the necessary input is derived from the states of the automata at neighbouring lattice-points. Thus the state of an cellular automata at time $t+1$ is a function of the states of the automata itself and its immediate neighbours at time $t$. All the finite automata in the lattice (group of cells) obey the same transition-table (rule table) and every cell changes his state at the same instant, time-step after time-step. CA's are a good example of the kind of computational paradigm sought after by Artificial Life: bottom-up, parallel, local determination of behaviour with minimal specification, and emerging complex phenomena from simple rules.

In order to study any natural phenomena, scientists are turning to a separation. A need to separate the notion of a formal specification of a machine (any that will reproduce the phenomena itself) – that is, a specification of the *logical structure* of the machine – from the notion of a formal specification of a machines's behaviour – that is, a specification of transitions that the machine will undergo. In general, we cannot derive behaviours from structure, nor can we derive structure from behaviours. So instead, in order to determine the behaviour of some machines and coupled phenomena, there is no recourse but to run them and see how they behave. This has consequences for the methods by which we (or nature) go about *generating* behaviour generators themselves, and from which any evolutionary and adaptive process seems to be essential. For instance, the most salient characteristic of living systems, from the behaviour generation point of view, is the *genotype/phenotype* distinction. The distinction is essentially one between a specification of machinery – the *genotype* – and the behaviour of that machinery – the *phenotype*.

The *genotype* is the complete set of genetic instructions encoded in the linear sequence of nucleotide bases that makes an organism's DNA. The *phenotype* is the physical organism itself – the structures that emerge in space and time as the result of the interpretation of the genotype of a particular environment. The process by which the phenotype develops through time under the direction of the genotype is called *morphogenesis*. Simulation plays an essential role in the study of *morphogenesis*. This was anticipated as early as 1952 by *Turing*, who wrote:

[…] The difficulties are such that one cannot hope to have any very embracing *theory* of such processes, beyond the statement of equations. It might be possible, however, to treat a few particular cases in detail with the aid of a digital computer. This method has the advantage that it is not so necessary to make simplifying assumptions as it is when doing a more theoretical type of analysis […]

What is notable is that this 1952 *Turing* words appears to have already the embedded features that characterise *bottom-up* approaches, in detriment of other kinds of approaches strictly *reductionist* (e.g. *top-down*). As an aside evidence, note the last *Turing* words on this sentence: it is not so necessary to make simplifying assumptions as it is when doing a more theoretical type of analysis […]. Visualisation itself, of simulation results facilitates their interpretation, and is used as a method for evaluating models. Lacking a formal measure of what makes two patterns or forms (such as trees) look alike (task that is, as we known, mainly related to the idea of perception), we rely on visual inspection comparing the models with the reality. Important however in this models, is that the natural and synthetic pigmentation patterns differ in details, yet we perceive them as fairly similar or familiar.

In *morphogenesis*, the individual genetic instructions are called *genes* and consist of short stretches of DNA. These instructions are *executed* (expressed) when their DNA sequence is used as a template for transcription. One may consider the genotype as a largely unordered 'bag' of instructions (a rule table, an alphabet, a group of primitives), each one of which is essentially the specification for a *machine* of some sort – passive or active. When instantiated, each such machine will enter into ongoing logical mechanisms, consisting largely of local interactions between other such machines. Each such instruction will be *executed* when its own triggering conditions are met and will have specific, local effects on structures in the cell (their neighborhood). Furthermore, each such instruction will operate within the context of all the other instructions that have been – or are being – executed.

The phenotype, then, consists of the structures and dynamics that emerge through time in the course of the execution of the parallel, distributed computation controlled by this genetic *bag* of instructions. Since genes interactions with one another are highly non-linear, the phenotype is a non-linear function of the genotype. As mentioned briefly above, the distinction between linear and non-linear systems is fundamental, and provides excellent insight into why the principles underlying the dynamics of life (or many other natural phenomena) should be so hard to find and understand. The simplest way to state the distinction is to say that linear systems are those for which the behaviour of the whole is just the sum of the behaviour of its parts, while for non-linear systems, the behaviour of the whole is more than the sum of its parts. Linear systems are those which obey the *principle of superposition*. We can break up complicated linear systems into simpler constituents parts, and analyse these parts *independently*. Once we have reached an understanding of the parts in isolation, we can achieve a full understanding of the whole system by *composing* our understandings of the isolated parts. This is the key feature of linear systems: by studying the parts in isolation we can learn everything we need to know about the complete system. Nature, however, is generally non-linear, where this type of approach is often impossible. Non-linear systems do *not* obey the principle of superposition. Even if we could break such systems up into simpler constituents parts, and even if we could reach a complete understanding of the parts in isolation, we would not be able to compose our understandings of the individual parts into an understanding of the whole system. The key feature of non-linear systems is that their primary behaviours of interest are properties of the interactions between parts, rather than being properties of the parts themselves, and these interaction-based properties necessarily disappear when the parts are studied independently. Analysis has not proved anywhere near as effective when applied to non-linear systems: the non-linear system must be treated as a whole. A different approach to the study of non-linear systems involves the inverse of analysis: *synthesis*. Rather than start with the behaviour of interest and attempting to analyse it into its constituent parts, we should start with constituent parts and put them together in the attempt to synthesize the behaviour of interest. Life, in the same way, is a property of *form*, not *matter*, a result of organization and re-organization of matter rather than something that inheres in the matter itself. Neither nucleotides nor amino acids nor any other carbon-chain molecule is alive – yet put them together in the right way, and the dynamic that emerges out of their interactions is what we call life. It is effects, not things, upon which life is based – life is a kind of behaviour, not a kind of stuff – and as such, it is constituted of simpler behaviours, not simpler stuff. Behaviours themselves can constitute the fundamental parts of non-linear systems – *virtual parts*, which depend on non-linear interactions between physical parts for their very existence. Isolate the physical parts and the virtual parts cease to exist. It is the virtual parts of living systems that Artificial Life is after, and synthesis is its primary methodological tool.

Artificial life, is attempting to develop a new computational paradigm based on the natural processes that support living organisms. That is, aLife uses insights from biology to explore the dynamics of interacting information structures, that is, aiming to know what key aspects exist in this information relations that can evolve the whole process. Artificial Life has not adopted the computational paradigm (limited as a technology of computation, as a sequence of orders) as its underlying and primary methodology of behaviour generation, nor does it attempt to explain life, or any other natural phenomena, as a kind of computer program. Computers are providing an alternative medium within which to attempt to synthesize life. Modern computer technology has resulted in machinery with tremendous potential for the creation of life *in silico*, and by far more important, a medium to understand how real life and nature are. Needless to say that, the present common fear of AI is proportional to his ignorance, as was true for many tools found useful in this century. As pointed by *James Martin* (the one that forecast the *Wired Society* back in 1977, while *Bill Gates* still banged out business correspondence on a typewriter): "People think that when computers become intelligent they will become intelligent like we are. Nothing could be more far from truth". Nevertheless, some common underlying features could be found, which seems to be pertinent in the study and understanding of any system (natural or artificial), independently of his type of support.

Computers, instead, provide (and should be viewed as) as an important laboratory tool for the study of life and many natural phenomena, as an alternative devoted exclusively to the incubation of information structures. The advantage of working with information structures is that information has no intrinsic size. The computer is the *tool* for the manipulation of information, whether that manipulation is a consequence of our actions or a consequence of the actions of the information structure themselves. Computers themselves will not be alive, rather they will support informational universes within which dynamic populations of informational 'molecules' (or *memes*, as proposed by *Dawkins*, as the cultural information genes, or vehicle, within one specific society) engage in informational 'biochemistry'. This view of computers as workstations for performing scientific experiments within artificial universes is fairly new, but is rapidly becoming accepted as a legitimate, even necessary, way of pursuing science. In the days before computers, scientists worked primarily with systems whose defining equations could be solved analytically, and ignored those whose defining equations could *not* be solved. This was the case, for instance, in many analytical systems trying to explain how the global weather changes, or trying to forecast the behaviour of a fire propagating in a specific terrain. As we now know, global weather is a chaotic non-linear system, where a flap of a butterfly wing in Peking can develop a huge storm in New-York, few days later. In the absence of analytical possible solutions, the equations would have to be integrated over and over again, essentially simulating the time behaviour of the system. Without computers to handle the mundane details of these calculations, such an undertaking was unthinkable except for the simplest cases. Given this mundane calculations to computers, the realm of numerical simulation is opened up for exploration. 'Exploration' is an appropriate term for the process, because the numerical simulation of systems allows one to explore the system's behaviour under a wide range of parameter settings and initial conditions.

The heuristic value of this experimentation cannot be overestimated. One often gains tremendous insight for the essential dynamics of a system by observing its behaviour under a wide range of initial conditions. Moreover, computers are beginning to provide scientists with a new paradigm for modeling the world. When dealing with essentially unsolvable governing equations, the primary reason for producing a formal mathematical model (the hope of reaching an analytic solution by symbolic manipulation) is lost. It has become possible, for example, to model turbulent flow in a fluid by simulating the motions of its constituent particles – not just approximating *changes* in concentrations of particles at particular points, but actually computing their motions exactly. The same is true for understanding and modeling people in overcrowded soccer stadiums, or for instance, in gaining insight on how traffic jams emerge, from very simple inner rules. Again, the best way to tackle it, is to look at the whole process (as within a helicopter !), synthesizing which basic and simple logical rules (generally independent from the phenomena itself) govern the multitude of parts, emerging a global and complex behaviour. What is essential in these type of systems, is not the parts and their intrinsic natures (at least strictly), but namely their relationships (among themselves and with their environment).

Within this same context, let us return again to genotype/phenotype distinction and on the possibility of the development of a behavioural phenotype. One paradigmatic model is the one of *Craig Reynolds*, who in 1987 has implemented a simulation of flocking behaviour. Now, if we think for a moment, none type of analytical differential equations was been able to tackle (or model) this type of natural phenomena. In the *Reynolds* model, however – which is meant to be a general platform for studying the qualitatively similar phenomena of flocking, herding and schooling – one has a large collection of autonomous but interacting objects (which *Reynolds* refer as *Boids*), inhabiting a common simulated environment.

The modeler can specify the manner in which the individual *Boids* will respond to *local* events or conditions. The global behaviour of the aggregate of *Boids* is strictly an emergent phenomena, where none of the rules for the individual *Boids* depends on global information, and the only updating of the global state is done on the basis of individual *Boids* responding to local conditions. Note that, the underlying system *nature* is similar in many ways to a *Cellular Automata*, mentioned earlier. Again, each *Boid* (*cell* for the CA) in the aggregate shares the same behavioural 'tendencies':

· To maintain a minimum distance from other objects in the environment, including other *Boids*,
· To match velocities with *Boids* in its neighbourhood, and
· To move towards the perceived centre of mass of the *Boids* in its neighbourhood.

These are the only rules governing the behaviour of the aggregate. These rules, then, constitute the generalized genotype of the *Boids* system. What is amazing, is that they say nothing about structure, or growth and development, or even about birds nature, but they determine the behaviour of a set of interacting autonomous objects, out of which very natural motion emerges.

With the right settings for the parameters of the system, a collection of *Boids* released at random positions within a volume will collect into a dynamic flock, which flies around environmental obstacles in a very fluid and natural manner, occasionally breaking up into sub-flocks as the flock flows around both sides of an obstacle. Once broken up into sub-flocks, the sub-flocks reorganize around their own, now distinct and isolated centre of mass, only to re-emerge into a single flock again when both (or more) sub-flocks emerge at the fair side of the obstacle and each sub-flock *feels* anew the mass of the other sub-flock.

The flocking behaviour itself constitutes the generalized phenotype of the *Boids* system. It bears the same relation to the genotype as an organism's morphological phenotype bears to its molecular genotype. The same distinction, between the specification of machinery and the behaviour of machinery is evident. Through development (or time), local rules governing simple non-linear interactions at the lowest level of complexity, emerge global behaviours and structures at the highest level of complexity. Finally, Artificial Life (as a truly interdisciplinary scientific field) may be viewed as an attempt to understand high-level behaviour from low-level rules, for example, on how the simple interactions between ants and their environment lead to complex trail-following behaviour. But by far more important than studying ants itself, is to study how they organize themselves, through out a simple adaptive *mechanism* that seems to be present in many natural phenomena of our world. An understanding of such relationships in particular systems can suggest novel solutions to complex real-world problems such as disease prevention, pattern recognition, stock-market prediction, or data mining on the Internet (to name up a few).

Other similar scientific models and ideas, can be found in *Natural Self-organising systems* (*Bilchev*). A great majority of natural and artificial systems are of complex nature, and scientists choose more often than not to work on systems simplified to a minimum number of components in order to observe *pure* effects. An alternative approach, often known as the *complex systems dynamics* approach (*Weisbuch*), is to simplify as much as possible the components of the system, so as to take into account their large number. This idea has emerged from a recent trend in research known as the *physics of disordered systems*, but can also be found earlier in many *aLife* models (mentioned above) that follow bottom-up approaches. As suggested by *Langton*, the key concept in *aLife* is emergent behaviour. Natural life emerges out of the organised interactions of a great number of nonliving molecules, with no global controller responsible for the behaviour of every part. Rather, every part is a behaviour itself, and life is the behaviour that emerges from out of all of the local interactions among individual behaviours. It is this bottom-up, distributed local determination that aLife employs in its primary methodological approach to the generation of lifelike behaviours. Again, this seems to happen in real systems like our own brain; in rapid succession, research has revealed that the brain uses discrete systems for different types of learning (*Damásio*).

*Complex dynamic systems* in general show interesting and desirable behaviours as *flexibility* (in vision or speech understanding tasks, the brain is able to cope with incorrect, ambiguous or distorted information, or even to deal with unforeseen or new situations without showing abrupt performance breakdown) or *versatibility* quoting *Dorigo* and *Colorni*, *robustness* (keep functioning even when some parts are locally damaged - *Damásio*), and they operate in a *massively parallel fashion*. As we know, systems of this kind abound in nature. A vivid example is provided by the behaviour of a society of termites (*Courtouis*). And, as a key feature, *complex dynamical systems* show and provide emergent properties. Again, this means that the behaviour of the system as a whole can no longer be viewed as a simple superposition of the individual behaviours of its elements, but rather as a side effect of their collective behaviour. Contained in this notion is the idea that properties are not *a priory* predictable from the structure of the local interactions and that they are of functional significance. The computation to be performed is contained in the dynamics of the system, which in turn is determined by the nature of the local interactions between the many elements.

Many of the dynamical computation systems that have been developed today find their equivalent in nature, and all of them shows, directly or not, important emergent properties (among other lifelike features). A non-extensive list of possible paradigmatic examples include, *Genetic Algorithms*, *Memetic Algorithms*, *Spin Glass Models*, *Connectionist Architectures* and *Artificial Neural Networks*, *Reaction-Diffusion* systems, *Self-Organizing Maps*, *Simulated Annealing* methods, *Artificial Imunne systems*, *Cellular Automata*, *L-Systems*, *Gradient Vector Flow* and *Snakes*, *Differential Evolution*, *Correlational Opponent Processing* and *Particle Swarm Optimization*. As an example, biological metaphors offer insight into many aspects of computer viruses and can inspire defenses against them. That is the case with some applications of *Immunological Computation*, and *Artificial Immune Systems*. The immune system is highly distributed, highly adaptive, maintains a memory of past encounters, is self-organising in nature and has the ability to continually learn about new encounters. From a computational viewpoint, the immune system has much to offer by way of inspiration. Detection of specific patterns in large databases is one possible application. Autonomous alert collision systems, in route management for airplanes is another.

Evolutionary computation is another example. In the spirit of *Von Neumann*, *John Holland* has attempted to abstract the logical form of the natural process of biological evolution in what is currently known as the *Genetic Algorithm* (GA). In the GA, a genotype is represented as a character string that encodes a potential solution to a problem. For instance, the character string (*chromosome*) might encode the weight matrix of a neural network, or the rule table of any *Cellular Automata*. These character strings are rendered as phenotypes via a problem-specific interpreter, which constructs, for example, the artificial neural network or the cellular automata machine specified by each genotype, evaluates its performance in the problem domain, and provides it with a specific fitness value. From this point the GA implements natural selection by making more copies of the character strings representing the better performing phenotypes. The GA generates variant genotypes by applying genetic operators to these character strings. The genetic operators typically consist of *reproduction*, *cross-over*, and *mutation*, with occasional usage of *inversion* and *duplication*. What is interesting is that "poor" individuals along several generations, often encode in parts of their genotypes, the key for the best solutions (artificial individuals) to become better. The best GA solution, is in some sense a product of the GA collective change of information, a product of the whole, being diversity a key aspect in the process, and a way for the artificial algorithm to balance his own exploration/exploitation duality character on the fitness landscape (space of possible solutions). Such evolutionary approaches are being applied to tasks such as optimisation, search procedures, classification, adaptation, among others.

Yet, another prominent example, are *Artificial Ant Systems*. In "*Godel, Escher, Bach*", *Douglas Hofstadter* explores the difference between an ant colony as a whole and the individual that compose it. According to *Hofstadter*, the behaviour of the whole colony is far more sophisticated and of very different character than the behaviour of the individual ants. A colony's collective behaviour exceeds the sum of its individual member's actions (so-called *emergence*) and is most easily observed when studying their foraging activity. Most species of ants forage collectively using chemical recruitment strategies, designated by *pheromone* trails, to lead their fellow nest-mates to food sources.

This analogy with the way that real and natural ant colonies work and migrate, has suggested the definition in 1991/92 of a new computational paradigm, which is called the *Ant System* (*Dorigo / Colorni*). In these studies (again) there is *no pre-commitment* to any particular representational scheme: the desired behaviour is specified, but there is minimal specification of the mechanism required to generate that behaviour, i.e. *global behaviour evolves from the many relations of multiple simple behaviours*. Since then several studies were conducted to apply this recent paradigm – or analogous ones - in real case problems, with successful results. The new heuristic has the following desirable characteristics: (1) It is *versatile*, in that it can be applied to similar versions of the same problem; (2) It is *Robust*. It can be applied with only minimal changes to other problems (e.g. combinatorial optimisation problems such as the quadratic assignment problem - QAP, travelling salesman problem - TSP, or the job-shop scheduling problem - JSP);…and (3) It is a *population based approach*. This last property is interesting since it allows the exploitation of positive feedback as a search mechanism (the collective behaviour that emerges is a form of *autocatalytic* "snow ball" - that reinforces itself - behaviour, where the more the ants follow a trail, the more attractive that trail becomes for being followed). It also makes the system amenable to parallel implementations (though, only the intrinsically parallel and distributed nature of these systems are generally considered).

An important feature in many of these dynamical computational systems is that of *interaction* (e.g. competition-cooperation duality). Cooperation involves a collection of agents – global behaviours, if we strictly follow *Langton* words - that interact by communicating information, or hints (usually concerning regions to avoid or likely to contain solutions) to each other while solving a problem. This duality interaction can also be found in the well known *Prisoner Dilemma* benchmark problem, used in many Evolutionary Algorithms including Genetic Algorithms. The information exchanged may be incorrect at times and should alter the behaviour of the agents receiving it. Another example of cooperative problem solving is the use of the Genetic Algorithm to find states of high fitness in some abstract space. In a Genetic Algorithm, members of a population of states exchange pieces of themselves or mutate to create new populations, often containing states of higher fitness. And yet another example are *Neural Networks*, where the output of one neuron affects the behaviour (or state – under the light of *Cellular Automata* theory) of the neuron receiving it, and so on. Reporting to the real nature and quoting *Damásio*, we are barely beginning to address the fact that interactions among many non-contiguous brain regions probably yield highly complex biological states that are vastly more than the sum of their parts. It is important, however, to point out that the brain and mind are not a monolith: they have multiple structural levels, and the highest of those levels creates instruments that permit the observation of the other levels.

On the other hand, it has become widely recognized that the past symbol-oriented community in AI only supported models in research that were far too rigid and specialized, focussing on well-defined problems that generally are rare to found in the real-world, that is, being too inflexible to function well outside the domains for which they were designed. Thus, they are often unable to deal with exceptions to rules, or to exploit fuzzy, approximate, or heuristic fragments of knowledge. We now know, that many of the "toy" problems of the past, has become the most difficult ones. Learning, recognition, adaptation, perception, visual capabilities in general, are among such examples. However, connectionist systems (the symbolic AI counterpart) seems to be in the right way, exploring the relations and capabilities among many simple parts, into the emergence of what we experience as a coherent and cognitive whole.

One paradigmatic case is that of perception. *Ramos* (CVRM / IST – Technical University of Lisbon) explored the application of Artificial Ant Systems into Pattern Recognition problems, namely to the sub-problem of image segmentation, i.e., to find homogeneous regions in any digital image, in order to extract and classify them. The application of these heuristics onto image segmentation looks very promising, since segmentation can be looked as a clustering and combinatorial problem, and the grey level image itself as a topographic map (where the image is the *ant colony playground*).

The distribution of the pheromone (a volatile and chemical substance) represents the memory of the recent history of the swarm, and in a sense it contain information which the individual ants are unable to hold or transmit. In this artificial system, there is no direct communication between the organisms but a type of indirect communication through the pheromonal field. In fact, ants are not allowed to have any memory and the individual's spatial knowledge is restricted to local information about the whole colony pheromone density. Particularly interesting for the present task, i.e. trying to evolve perceptive capabilities, the self-organisation of ants into a swarm and the self-organisation of neurones into a brain-like structure are similar in many respects (*Chialvo*, *Millonas*). Swarms of social insects construct trails and networks of regular traffic via a process of pheromone laying and following. These patterns constitute what is known in brain science as a *cognitive map*. The main differences lies in the fact that insects write their spatial memories in the environment, while the mammalian cognitive map lies inside the brain, a fact that also constitutes an important advantage in the present model. As mentioned by *Chialvo*, this analogy can be more than a poetic image, and can be further justified by a direct comparison with the neural processes associated with the construction of cognitive maps in the hippocampus. *Wilson*, for instance, forecasted the eventual appearance of what he called "a stochastic theory of mass behaviour" and asserted that "the reconstruction of mass behaviours from the behaviours of single colony members is the central problem of insect sociobiology". He forecasted that our understanding of individual insect behaviour together with the sophistication with which we will able to analyse their collective interaction would advance to the point were we would one day posses a detailed, even quantitative, understanding of how individual "probability matrices" would lead to mass action at the level of the colony. By replacing *colony members* with *neurones*, *mass behaviours* or *colony* by *brain behaviour*, and *insect sociobiology* with *brain science* the above paragraph could describe the paradigm shifts in the last twenty-five years of progress in the brain sciences.

Also a key issue is that, perception itself, as a human feature is being modelled and analysed by *Gestalt* psychology and philosophical systems since, at least 1910 (*Wertheimer*). It is of much interest to follow that this kind of scientific works point out that perception is a product of a synergistic whole effect, i.e. the effect of perception is generated not so much by its individual elements (e.g. human neurones) as by their dynamic interrelation (collective behaviour) – phenomena that can be found easily in many computational paradigms briefly described above, or even in *Neural Network* computational models, where data generalisation, *N* dimensional matrix re-mapping, pattern classification or forecasting abilities are known to be possible. As putted by *Limin Fu* in his own words, the *intelligence* of a Neural Network emerges from the collective behaviour of neurones, each of which performs only very limited operations. Even though each individual neuron works slowly, they can still quickly find a solution by working in parallel. This fact can explain why humans can recognize a visual scene faster than a digital computer, while an individual brain cell responds much more slowly than a digital cell in a VLSI (*Very Large Scale Integration*) circuit. Also, this *brain metaphor* suggests how to build an intelligent system which can tolerate faults (fault tolerance) by distributing information redundantly. It would be easier to build a large system in which most of the components work correctly than to build a smaller system in which all components are perfect. Another feature exhibited by the brain is the associative type of memory. The brain naturally associates one thing with another. It can access information based on contents rather than on sequential addresses as in the normal digital computer. The associative, or content-addressable, memory accounts for fast information retrieval and permits partial or approximate matching. The brain seems to be good at managing fuzzy information because of the way its knowledge is represented.

Typically these systems form a structure, configuration, or pattern of physical, biological, sociological, or psychological phenomena, so integrated as to constitute a functional unit with properties not derivable from its parts in summation (i.e. non-linear) – *Gestalt* in one word (*Krippendorff*) (the English word more similar is perhaps *system*, *configuration* or *whole*). This synergetic view, derives from the holistic conviction that the whole is more than the sum of its parts and, since the *energy* in a whole cannot exceed the sum of the energies invested in each of its parts (e.g. first law of thermodynamics), that there must therefore be some quantity with respect to which the whole differs from the mere aggregate. This quantity is called synergy and in many aLife computational systems can be seen as their inherent emergent and *autocatalytic* properties (process well known in many *Reinforcement Learning* models, namely in Q-learning methods often used in autonomous-agents design (*Mitchell* / *Maes*).

Part of what we know see in the MC$^2$ UTOPIA Biennial Art Exposition, was due to a model that has explored the application of these features into digital images, replacing the normal colony *habitat*, by grey levels, extending the capabilities of pheromone deposition into different situations, allowing a process of perceptual morphogenesis. In other words, from local and simple interactions to global and flexible adaptive perception. In those experiments, the emergence of network pheromone trails, for instance, are the *product* of several simple and local interactions that can evolve to complex patterns, which in some sense translate a meta-behaviour of that swarm. Moreover, the translation of one kind of low-level structure of information (present in a large number) to one meta-level is minimal. Although that behaviour is specified (and somehow constrained), there is minimal specification of the mechanism required to generate that behaviour; global behaviour evolves from the many relations of multiple simple behaviours, without global coordination, and using indirect communication (through the environment). One paradigmatic and abstract example is the notion, within a specified population, of *common-sense*, being the meta-result a type of *collective-conscience*. Needless to say, that some features are acquired (through out the evolving relation with the habitat), being others inner components of each part. Though, what is interesting to note is that we do not need to specify them. Moreover, the present model shows important adaptive capabilities, as in the presence of sudden changes in the *habitat*.

Even if the model is able to quickly adapts to one specific environment, evolving from one empty pheromonal field, *habitat* transitions point that, the whole system is able to have some memory from past environments (i.e. convergence is more difficult after *learning* and *perceiving* one *habitat*). This emerged feature of *résistance*, is somewhat present in many of the natural phenomena that we find today in our society.

Another common feature found in many of these computational paradigms, is that of *synergy*. Synergy (from the Greek word *synergos*), broadly defined, refers to combined or co-operative effects produced by two or more elements (parts or individuals). The definition is often associated with the quote "the whole is greater than the sum of its parts" (Aristotle, in *Metaphysics*), even if it is more accurate to say that the functional effects produced by wholes are different from what the parts can produce alone. Synergy is a ubiquitous phenomena in nature and human societies alike. One well know example is provided by the emergence of self-organization in social insects, via direct (mandibular, antennation, chemical or visual contact, etc) or indirect interactions. The latter types are more subtle and defined by *Grassé* as *stigmergy* to explain task coordination and regulation in the context of nest reconstruction in *Macrotermes* termites. An example, could be provided by two individuals, who interact indirectly when one of them modifies the environment and the other responds to the new environment at a later time. In other words, *stigmergy* could be defined as a typical case of environmental synergy. *Grassé* showed that the coordination and regulation of building activities do not depend on the workers themselves but are mainly achieved by the nest structure: a stimulating configuration triggers the response of a termite worker, transforming the configuration into another configuration that may trigger in turn another (possibly different) action performed by the same termite or any other worker in the colony. Another illustration of how *stimergy* and self-organization can be combined into more subtle adaptive behaviors is recruitment in social insects. Self-organized trail laying by individual ants is a way of modifying the environment to communicate with nest mates that follow such trails. It appears that task performance by some workers decreases the need for more task performance: for instance, nest cleaning by some workers reduces the need for nest cleaning. Therefore, nest mates communicate to other nest mates by modifying the environment (cleaning the nest), and nest mates respond to the modified environment (by not engaging in nest cleaning); that is *stigmergy*.

In other words, perception and action only by themselves can evolve adaptive and flexible problem-solving mechanisms, or emerge communication among many parts. The whole and their relationships (that is, the next layer in complexity) emerges from the relationship of many parts, even if these latter are acting strictly within and according to any sub-level of basic and simple strategies, *ad-infinitum* repeated. Quoting *Einstein*, the system "should be made as simple as possible, but not simpler". Division of labor is another paradigmatic phenomena of *stigmergy*. Simultaneous task performance (parallelism) by specialized workers is believed to be more efficient than sequential task performance by unspecialized workers. Parallelism avoids task switching, which costs energy and time. A key feature of division of labor is its plasticity. Division of labor is rarely rigid. The ratios of workers performing the different tasks that maintain the colony's viability and reproductive success can vary in response to internal perturbations or external challenges.

But by far more crucial to the design of any "Machine of Collective Conscience", is how ants form piles of items such as dead bodies (corpses), larvae, or grains of sand. There again, *stigmergy* is at work: ants deposit items at initially random locations. When other ants perceive deposited items, they are stimulated to deposit items next to them, being this type of cemetery clustering organization and brood sorting a type of self-organization and adaptive behavior. *Théraulaz* and *Bonabeau* described for instance, a model of nest building in wasps, in which wasp-like agents are stimulated to deposit bricks when they encounter specific configurations of bricks: depositing a brick modifies the environment and hence the stimulatory field of other agents. These asynchronous automata (designed by an ensemble of algorithms) move in a 3D discrete space and behave locally in space and time on a pure stimulus-response basis. There are other types of examples (e.g. prey collectively transport), yet *stimergy* is also present: ants change the perceived environment of other ants (their cognitive map, according to *Chialvo* and *Millonas*), and in every example, the environment serves as medium of communication.

What all these examples have in common is that they show how *stigmergy* can easily be made operational. As mentioned by *Bonabeau*, that is a promising first step to design groups of artificial agents which solve problems: replacing coordination (and possible some hierarchy) through direct communications by indirect interactions is appealing if one wishes to design simple agents and reduce communication among agents. Another feature shared by several of the examples is incremental construction: for instance, termites make use of what other termites have constructed to contribute their own piece. In the context of optimization (though not used directly on the present work), incremental improvement is widely used: a new solution is constructed from previous solutions (see ACO paradigm, *Dorigo* et al). Finally, *stigmergy* is often associated with flexibility: when the environment changes because of an external perturbation, the insects respond *appropriately* to that perturbation, as if it were a modification of the environment caused by the colony's activities. In other words, the colony can collectively respond to the perturbation with individuals exhibiting the same behavior. When it comes to artificial agents, this type of flexibility is priceless: it means that the agents can respond to a perturbation without being reprogrammed to deal with that particular instability. If we wish, for instance, to design a data mining classification system, this means that no classifier re-training is needed for any new sets of data-item types (new classes) arriving to the system, as is necessary in many classical models, or even in some recent ones. Moreover, the data-items that were used for supervised purposes in early stages in the colony evolution in his exploration of the search-space, can now, along with new items, be re-arranged in more optimal ways. Classification and/or data retrieval remains the same, but the system organizes itself in order to deal with new classes, or even new sub-classes. This task can be performed in real time, and in robust ways due to system's redundancy. Many experiments are now under their way at the CVRM-IST Lab (for instance, real-time marble and granite image classification, image and data retrieval, etc), along with the application of Genetic Algorithms, Neural Networks, and many others (based strictly on natural computation paradigms) to many problems in Natural Resources Management like forecasting water quality and control on river networks.

Recently, several scientific papers have highlighted the efficiency of stochastic approaches based on ant colonies for problem solving. This concerns for instance combinatorial optimization problems like the *Traveling Salesman* problem, the *Quadratic Assignment* problem, Routing problems, the Bin Packing problem, or Time Tabling scheduling problems. Numerical optimization problems have been tackled also with artificial ants, as well as Robotics.

Data and information clustering is also one of those problems in which real ants can suggest very interesting heuristics for computer scientists, and it is in fact a classic strategy often used in Image and Signal Processing. Examples can be found using for instance the classic *K*-Means Clustering model, or in extensions of it as in the well know ISODATA, or even with other hybrid and more recent approaches using Genetic Algorithms within image classification (*Ramos*). One of the first studies using the metaphor of ant colonies related to the above clustering domain is due to *Deneubourg*, where a population of ant-like agents randomly moving onto a 2D grid are allowed to move basic objects so as to cluster them. This method was then further generalized by *Lumer* and *Faieta*, applying it to exploratory data analysis, for the first time. In 1995, the two authors were then beyond the simple example, and applied their algorithm to interactive exploratory database analysis, where a human observer can probe the contents of each represented point (sample, image, item) and alter the characteristics of the clusters. They showed that their model provides a way of exploring complex information spaces, such as document or relational databases, because it allows information access based on exploration from various perspectives. However, this last work was never published due to commercial applications.

Following the emergent features described above, the idea behind the $MC^2$ Machine is simple to transpose for the first time, the mammalian cognitive map, to a environmental (spatial) one, allowing the recognition of what happens when a group of individuals (humans) try to organize different abstract concepts (words) in one habitat (via internet). Even if each of them is working alone in a particular sub-space of that "concept" *habitat*, simply rearranging notions at their own will, mapping "Sameness" into "Neighbourness", not recognizing the whole process occurring simultaneously on their society, a global collective-conscience emerges. Clusters of abstract notions emerge, exposing groups of similarity among the different concepts. The $MC^2$ machine is then like a mirror of what happens inside the brain of multiple individuals trying to impose their own conscience onto the group.

Take any swarm. Take any collective natural system, where many parts are present. Study it. Identify which rules are prominent at local neighbours. The simpler the better. Understand if they are similar in any other natural system. You will probably be astonished. Now, collect them together in any computer. Mix them. Play it and let them evolve by their own. Soon, you will perceive organization. Any type of organization. What you will see is nothing more than the decay of entropy. But, don't stop it and feed the system with diversity. Re-inject knowledge if you think they will take profit of it. Memory among the whole is emerging. Even better than that: parts of the system at different locations can perceive the whole. Now, from time to time, allow the system to become slightly chaotic. Evaporation is one way. Oh, yes! Solutions found so far become more robust and flexible. Now, take this whole as a unit. And take any other whole. And another one. Take a lot of wholes and collect them in a computer, or in any other type of information structure. Put them in another layer of complexity. Mix them. Play it and let them evolve..., are you pleased?

Take a swarm. Play it. Let them evolve trails of pheromone in any 2D grid. Now, replace the grid by any digital image. Use this image as the colony Habitat. As their playground. Wait. Be patient. Then, take that image and put another one. Have you seen what happened? They have some memory of the past, isn't it?! Memory of what? Now take one map of pheromone at any time step. Use it, as it was any type of structure of information. Grab it. See it as the swarm cognitive map. Forget this 2D draw. Evolve the third dimension, that is, the frequency of pheromone at any cell of the Habitat. Join the edges, since they move in a toroidal space. Now, replay the system, seeing only this. Enjoy it! You are now in another layer of complexity. You are seeing the brain of the collective whole, evolving on time.

Now, forget ants. Imagine a 2D environment, with many cells. Imagine that some cells have individuals. Human individuals. And, imagine that in some other cells, objects are present. These objects are letters. These objects can be moved by the individuals. Allow the system to have pheromone deposition. Allow also, evaporation from time to time. Launch this *habitat* and these possibilities onto any world web site,… and wait. If we free the system, letters will re-organize themselves onto words. Many words, in many languages. Words, then, will evolve into concepts. Concepts that reflect the common thought between all the virtual society. A society composed of many autonomous individuals, which in a non-hierarchical way emerge something that exceeds the simple summation of their positions, and trying to force their views. Can you imagine what type of cognitive map will appear? What is then, a *pheromone* trail, within this specific context? Have you imagined how useful this can be?!

Through a Internet site reflecting the "words habitat", the users (humans) choose, gather and reorganize some types of letters, words and concepts. The overall movements of these letter-objects are then mapped into a public space. Along this process, two shifts emerge: the virtual becomes the reality, and the personal subjective and disperse beliefs become onto a social and politically significant element.

The MC$^2$ machine will reveal then what happens in many real world situations; cooperation among individuals, altruism, egoism, radicalism, and also the resistance to that radicalism, memory of that society on some extreme positions on time, but the inevitable disappearance of that positions, to give rise to the convergence to the group majority thought (Common-sense?), eliminating good or bad relations found so far, among in our case, words and abstract notions. Even though the machine composed of many human-parts will "work" within this restrict context, she will reveal how some relationships among notions in our society (ideas) are only possible to be found, when and only when, simple ones are found first (the minimum layer of complexity), neglecting possible big steps of a minority group of visionary individuals. What effect these local jumps of complexity have in the long run? And, if really the information process evolution of the system (giving jumps at several complexity levels) follows (from the inner basic to the global complex) the conceptual space of: Rules ? Relations ? Behaviours; what intermediate relations are really significant, allowing the system to emerge a coherent and cognitive Whole. Is there (in our society) any need for a critical mass of knowledge, in order to achieve other layers of complexity? Roughly, she will reveal for instance how democracies can evolve and die on time, as many things in our impermanent world.

We believe, that what we are showing here today at the UTOPIA Biennial Art Exposition, is a possible walk, up to *Life-like Complexity and Behaviour*, from bottom, basic and simple bio-inspired heuristics – a walk, up into the *morphogenesis* of information. We believe that these are the first steps into the design of truly collective, flexible, cognitive and adaptive forms of information structures, whatever they may be, or whatever they may represent, among many possible and specific contexts. From the construction of collective non-hierarchical and permanently evolving texts, to the emergence of self-organising and distributive *Local Area Networks* of concepts (through the use of flexible self-reinforcing links, as *pheromone* trails), applications are immense. Can we, for instance, induce new forms of Art?

Finally, we would like to quote *Pierre Lévy* (in, *World Philosophy*, Éditions Jacob, 2000):

[…] Institutions, States, Public Administrations, Universities, Museums, Enterprises, Associations, Groups of Interest, Individuals, all those neglecting the study towards the best way to introduce themselves in the processes of collective intelligence (that is, the social processes of interchange and knowledge production) and distributive ways of attention, that evolve under a planetary cyberspace, will miss the opportunity of achieving the slightest role on the world of the future […]